\documentclass[11pt]{article}

\usepackage[final]{custom}

\usepackage{times}
\usepackage{latexsym}

\usepackage[T1]{fontenc}

\usepackage[utf8]{inputenc}

\usepackage{microtype}

\usepackage{inconsolata}

\usepackage{graphicx}

\usepackage{subcaption}

\title{When 2D Tasks Meet 1D Serialization: \\On Serialization Friction in Structured Tasks}

\author{First Author \\
  Affiliation / Address line 1 \\
  Affiliation / Address line 2 \\
  Affiliation / Address line 3 \\
  \texttt{email@domain} \\\And
  Second Author \\
  Affiliation / Address line 1 \\
  Affiliation / Address line 2 \\
  Affiliation / Address line 3 \\
  \texttt{email@domain} \\}

\author{Chung-Hsiang Lo$^{1}$\thanks{Co-first authors.} , Lu Li$^{2*}$, Diji Yang$^{3}$\thanks{Co-project leaders.} , Tianyu Zhang$^{4,5\dagger}$, Yunkai Zhang$^{6}$,\\ \textbf{Yoshua Bengio$^{4,5}$,} \textbf{Yi Zhang$^{3}$} \\ \\  
$^{1}$Northeastern University \quad
$^{2}$University of Pennsylvania \quad
$^{3}$UC Santa Cruz \\
$^{4}$Mila - Quebec AI Institute \quad
$^{5}$University of Montreal \quad
$^{6}$BAIR, UC Berkeley \\ 
\texttt{lo.chun@northeastern.edu}, \quad
\texttt{luli1@sas.upenn.edu}, \quad
\texttt{\{dyang39, yiz\}@ucsc.edu} \\
\texttt{\{tianyu.zhang, yoshua.bengio\}@mila.quebec}, \quad 
\texttt{yunkai\_zhang@berkeley.edu}
}

\begin{document}
\maketitle
\begin{abstract}


In the LLM era, many symbolic and structured problems are presented to models through 1D text serialization. Yet some such problems are natively two-dimensional: their relevant relations, such as row--column correspondence or spatial adjacency, are defined by position in a 2D layout rather than by sequential order. This raises a representational question: does preserving the same symbolic entries in a 1D sequence also preserve the relational structure needed for computation? We study this issue through the lens of \textit{serialization friction}: the representational mismatch in which the same underlying task instances and entries are still present, but relations that depend on layout become implicit under 1D serialization. 
The study uses a controlled synthetic testbed of three tasks: matrix transpose, Conway's Game of Life, and LU decomposition. In each task, the same instances are presented either as 1D text serialization or as their native 2D layout rendered as an image. Across this testbed, 1D serialization degrades more sharply as task size grows, and errors under serialization exhibit spatially structured patterns, suggesting that this presentation choice is consequential within our testbed. 
To further interpret these results, we add supplementary analyses that include a within-visual probe and an additional comparison of the two input presentations under the mixed-training transpose setting.  These findings suggest that, for layout-defined tasks, reducing inputs to 1D serialization is not a neutral choice of representation.

\end{abstract}

\section{Introduction}

Large language models (LLMs) have achieved remarkable progress on text-based reasoning tasks~\citep{brown2020gpt3}, yet their textual interface still exposes inputs through a 1D token sequence representation~\citep{vaswani2017attention}. This interface fits ordinary text, but some structured tasks organize their relevant relations in 2D layouts rather than linguistic sequences. When such tasks are serialized into text, their symbolic content may remain present, but their positional and spatial relations become less directly expressed. This raises our central question: for tasks whose relevant structure is naturally expressed in 2D layout, is serializing them into 1D text a neutral interface choice?

We use \textit{serialization friction} to describe the representational mismatch that can arise when a task with native 2D structure is expressed through 1D text serialization. By structured 2D tasks, we mean problems whose native representation is a two-dimensional layout, where task-relevant relations such as row–column correspondence or spatial adjacency are defined by position rather than by sequential order. The issue is not necessarily that the symbolic content is removed: a serialized matrix or grid can still contain the same entries as its 2D counterpart. Rather, task-relevant relations that are explicit in the original layout, such as column alignment or spatial adjacency, become implicit under textual serialization and are no longer presented in their native layout form.


This question is related to, but distinct from, prior work on tables, documents, and layout-aware modeling~\citep{herzig2020tapas,huang2022layoutlmv3,lee2022pix2struct,sui2024table,fatemi2023talk,yin-etal-2025-talk,dantart2026topo0rag0}. Existing studies have shown that layout can be useful for particular data formats and applications, such as table understanding, document parsing, and visually grounded text processing~\citep{wei2025deepseekocr,cheng2025glyph,zhao2025vtcbench}. Our focus is different. We ask whether the common LLM-era reduction of a layout-defined symbolic instance to a 1D text sequence is itself a neutral interface choice. In other words, the object of study is not a new table or document benchmark, but the broader assumption that text-serializable structure can be safely treated as sequential text.

To examine this assumption, we construct a controlled synthetic testbed of three tasks that serve as witnesses rather than end applications: matrix transpose, Conway’s Game of Life~\citep{gardner1970fantastic}, and LU decomposition. These tasks are intentionally simple and programmatically generated, because their role is to isolate the representational question under exact supervision and evaluation. They expose three distinct forms of layout-defined computation: row–column alignment in matrix transpose, local spatial adjacency in Conway’s Game of Life, and sustained row–column structure across dependent operations in LU decomposition. We do not claim that these tasks represent all text-expressible problems, nor that 2D representations are generally superior to text. Instead, they provide controlled evidence for a claim: there exist text-serializable tasks for which reducing the input to a 1D sequence is not representation-neutral.

Across this testbed, we observe a consistent empirical pattern. When the same problem instances are flattened into 1D text, performance often degrades more sharply as task size grows, even though the symbolic entries are preserved. This degradation is not random: errors under serialization show spatial patterns tied to the structure of each task. This pattern is notable because recent work has found that multimodal models can struggle when text is merely visualized as pixels rather than provided directly as text~\citep{sun2026reading0,liu2026vista0bench0}. To distinguish 2D layout from image input alone, we further include a limited within-visual comparison in which the same 1D serialized string is rendered as an image. In this probe, the disrupted visual representation performs worse than the native 2D image layout, suggesting that rendering a 1D serialization as an image does not recover the performance of the native 2D representation. Taken together, these results are consistent with the serialization friction lens and suggest that tasks whose relevant relations are defined by 2D layout may warrant reconsideration when automatically reduced to 1D text serialization.

In summary, our contributions are as follows:
\begin{itemize} 

    \item We formulate \textit{serialization friction} as a representation-level lens for text-serializable but layout-defined symbolic tasks, where entries remain present under 1D serialization, but task-relevant relations become implicit.

    \item We build a controlled testbed of three tasks with native 2D structure: matrix transpose, Conway's Game of Life, and LU decomposition. The testbed lets us compare two practical ways of presenting the same underlying instances: flattening them into serialized text and preserving their native 2D layout as images.


    \item We provide diagnostic evidence that, for tasks in this class, 1D serialization is not a representation-neutral default: within our testbed, it is consistently associated with sharper degradation and task-structured error patterns, suggesting that originally 2D tasks should not be automatically reduced to 1D text by default.
    
\end{itemize}

\section{Related work}

\paragraph{Visual-Text Compression for Context Scaling.}
A recent alternative to token-based long-context scaling is \emph{visual-text compression}, which renders long sequences into images and relies on a vision encoder to produce a compact representation for downstream decoding. Glyph exemplifies this direction by treating long-context modeling as a multimodal problem and showing substantial compression while maintaining accuracy on long-context benchmarks, without requiring task-specific fine-tuning for each downstream use case~\citep{cheng2025glyph}. DeepSeek-OCR similarly studies optical 2D mapping as a compression mechanism and demonstrates that high compression ratios can still preserve strong decoding accuracy, reinforcing that vision token budgets can be traded against text length in a controllable way~\citep{wei2025deepseekocr}. In our work, rendered inputs serve as an operational probe for preserving native 2D structure.

\paragraph{Layout-Aware Modeling and 2D Inductive Biases.}
Prior work in Document AI and structured input modeling suggests that preserving 2D structure can improve understanding by making alignment, locality, and grouping explicit. Work on table understanding highlights the importance of representation for row--column structure~\citep{herzig2020tapas,wang2021tuta,yin2020tabert,sui2024table}, while layout-aware and OCR-free approaches such as LayoutLMv3, Donut, and Pix2Struct further show the value of exploiting page structure end to end~\citep{huang2022layoutlmv3,kim2021donut,lee2022pix2struct}. Related work also suggests that explicit visual structure and serialization choices can affect reasoning over structured inputs, including in vision-language and graph settings~\citep{izadi2025visualstructures,wei2024gita,fatemi2023talk,yin-etal-2025-talk}. Our setting differs in goal: rather than optimizing for document or table understanding, we use controlled structured tasks to examine what happens to originally 2D problems when they are presented through 1D text serialization.

\begin{figure*}[t]
    \centering
    \includegraphics[width=1.0\textwidth]{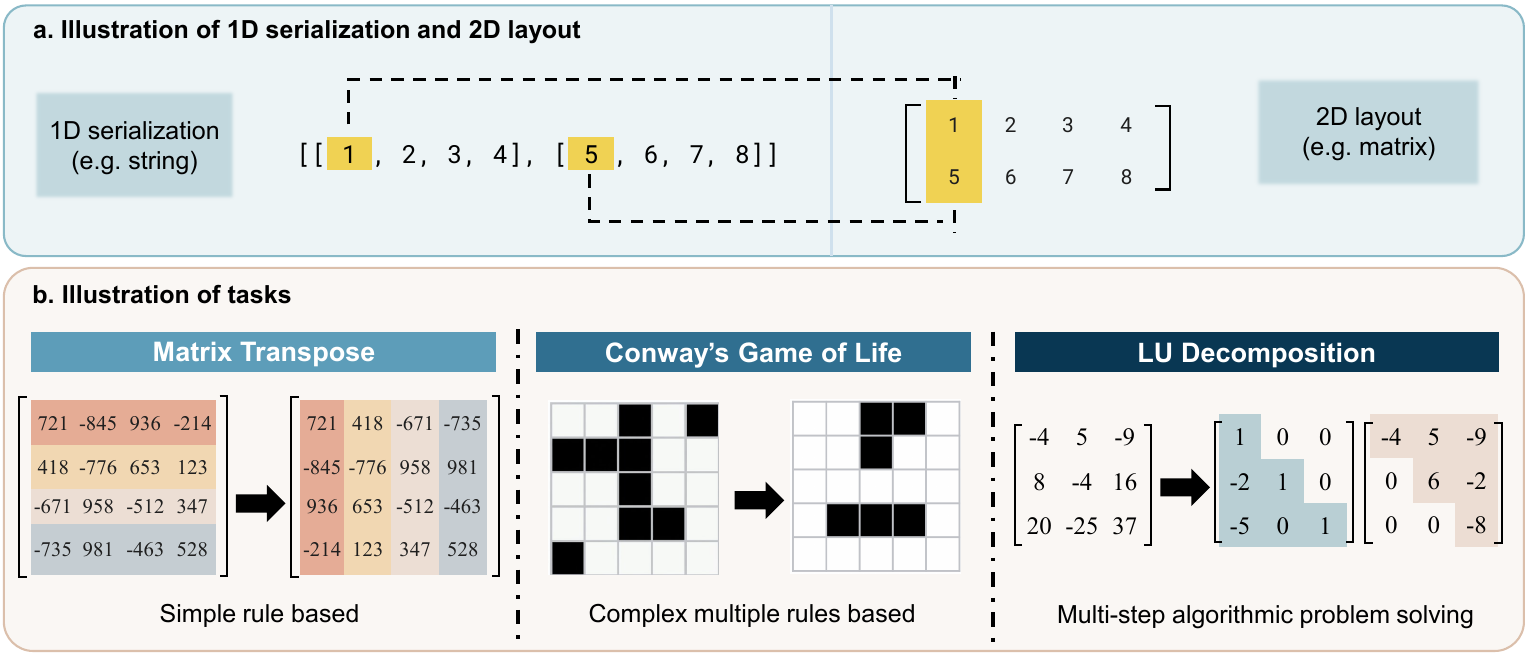}
    \caption{\textbf{a}. Illustration of serialization friction. In 2D layout, structural relations such as column alignment are explicit; under 1D serialization, the same relations are no longer explicitly encoded, but become implicit in sequential position and delimiters. \textbf{b}. Illustration of the three tasks used in our study: (i) matrix transpose, (ii) Conway's Game of Life, and (iii) LU decomposition. Details of the actual rendered inputs are provided in Appendix~\ref{app:rendering_details}. }
    \label{fig:friction_example}
\end{figure*}

\section{Concept}

In this section, we first define serialization friction, the representational mismatch that can arise when inherently 2D tasks are flattened into sequential form, and then describe the testbed through which we examine what happens to tasks with explicit 2D structure when they are presented through 1D text serialization.

\subsection{Serialization Friction in Structured 2D Tasks}


Concretely, serialization friction arises from a structural property of the text interface itself: common textual serializations reduce a layout-defined input to a 1D token sequence, making layout-dependent relations implicit in sequential position, delimiters, and tokenized content. This framing motivates the central comparison of our study: comparing textual serialization with a visual representation that preserves the native 2D layout on tasks where task-relevant structure is part of the computation.

\subsection{Representation Comparison Setup}

We instantiate this comparison as a system-level contrast between two practical input pathways, serialized 1D text and the native 2D layout rendered as an image, applied to the same family of tasks with explicit 2D structure. For brevity, we refer to these as the text condition and the visual condition throughout. The aim of this setup is not to account for every difference between the two conditions, but to focus on the presentation choice: we therefore hold fixed the underlying task instances, task definitions, prompts, train/eval splits, and evaluation criteria across both conditions.

\subsubsection{Input Representations}

We compare two ways of presenting the same underlying structured problem. In the \textbf{textual representation}, the input is serialized into a one-dimensional token sequence. In the \textbf{visual representation}, the same structured content is rendered in its original two-dimensional layout. 

To instantiate this comparison, we use GLM-4-9B-0414~\citep{glm2024chatglm} for text condition and Glyph~\citep{cheng2025glyph} for visual condition. Glyph is a visual-text compression method for long context understanding, which connects GLM with a vision encoder. This makes the GLM/Glyph pair a comparatively close match for studying how behavior changes across the two input forms. Concretely, GLM receives serialized text directly, whereas Glyph receives rendered 2D inputs through its visual front end.

\subsubsection{Task Suite}
\label{sec:task_suite}

We study three tasks whose computation depends on explicit 2D structure: matrix transpose, Conway's Game of Life \citep{gardner1970fantastic}, and LU decomposition. Full task definitions are provided in Appendix~\ref{app:task_definitions}.

\paragraph{Matrix Transpose.}
This setting is governed by explicit row--column correspondence in a 2D layout. The required transformation preserves all values and rearranges them solely according to the transposed structure, without introducing additional arithmetic or rule execution. It lets us examine whether 1D serialization is consequential for a task whose output follows directly from row--column correspondence.

\paragraph{Conway's Game of Life.} 
Here, correctness is determined by local neighborhood interaction on a 2D grid. The next state is produced by applying the same update rule to each cell based on its surrounding neighbors, so the relevant structure lies in spatial adjacency rather than row--column alignment alone. This task remains a one-step structure-driven prediction problem: the output is determined directly from the current grid configuration rather than through a sequence of dependent intermediate transformations. It lets us probe whether the same qualitative difference appears for rule-based computation over local 2D structure. 

\paragraph{LU Decomposition.}
In this setting, the relevant structure must be maintained across a sequence of dependent row--column operations. Producing a valid factorization requires the matrix to be progressively transformed while preserving relationships that constrain later steps of the computation. It lets us test whether the representation difference remains visible when 2D structure stays relevant throughout a sustained transformation process.

The first two tasks probe whether 1D serialization is consequential when the target transformation is directly tied to spatial alignment or local neighborhood interaction. LU decomposition instead examines a different role of structure, where relevant row--column relationships must be maintained and reused across a sequence of dependent operations.

Evaluation is task-specific but shared across the two conditions within each task. For matrix transpose and Conway's Game of Life, predictions are evaluated against the ground-truth. For LU decomposition, correctness is determined by whether the predicted lower and upper triangular factors satisfy the required structural constraints and can reconstruct the original matrix. Full dataset construction details, hyperparameter settings, and implementation specifics are provided in Appendix~\ref{app:implementation_details}.

\subsubsection{Instantiation Overview}

To support a controlled comparison, all tasks are instantiated from synthetic datasets with programmatically generated inputs and outputs. This design lets us vary task dimensions systematically, generate exact supervision targets, and ensure that the textual and visual conditions are derived from the same underlying task instances. 

We use different training paradigms according to the natural supervision signal of each task. For matrix transpose and Conway's Game of Life, we adopt supervised fine-tuning, since both tasks admit direct target outputs and primarily serve to test whether the two conditions behave differently under matched explicit supervision. For LU decomposition, we instead use reinforcement learning with verifiable reward, following recent work showing that outcome-verified optimization can effectively induce reasoning improvements in domains with objective correctness checks~\citep{deepseek-ai2025deepseek0r10}. Multiple valid intermediate trajectories may lead to a correct decomposition, and success depends on maintaining consistency across a sequence of dependent elimination steps. RLVR is therefore a natural fit for optimizing the task's outcome-level correctness criterion in this setting~\citep{yang2025depth0breadth,shao2024deepseekmath,yu2025dapo}. Because the optimization paradigm differs across tasks, we compare conditions within each task rather than treating absolute accuracies across tasks as directly comparable.


\begin{figure}[t]
    \centering
    \begin{subfigure}[t]{\linewidth}
        \centering
        \includegraphics[width=\linewidth]{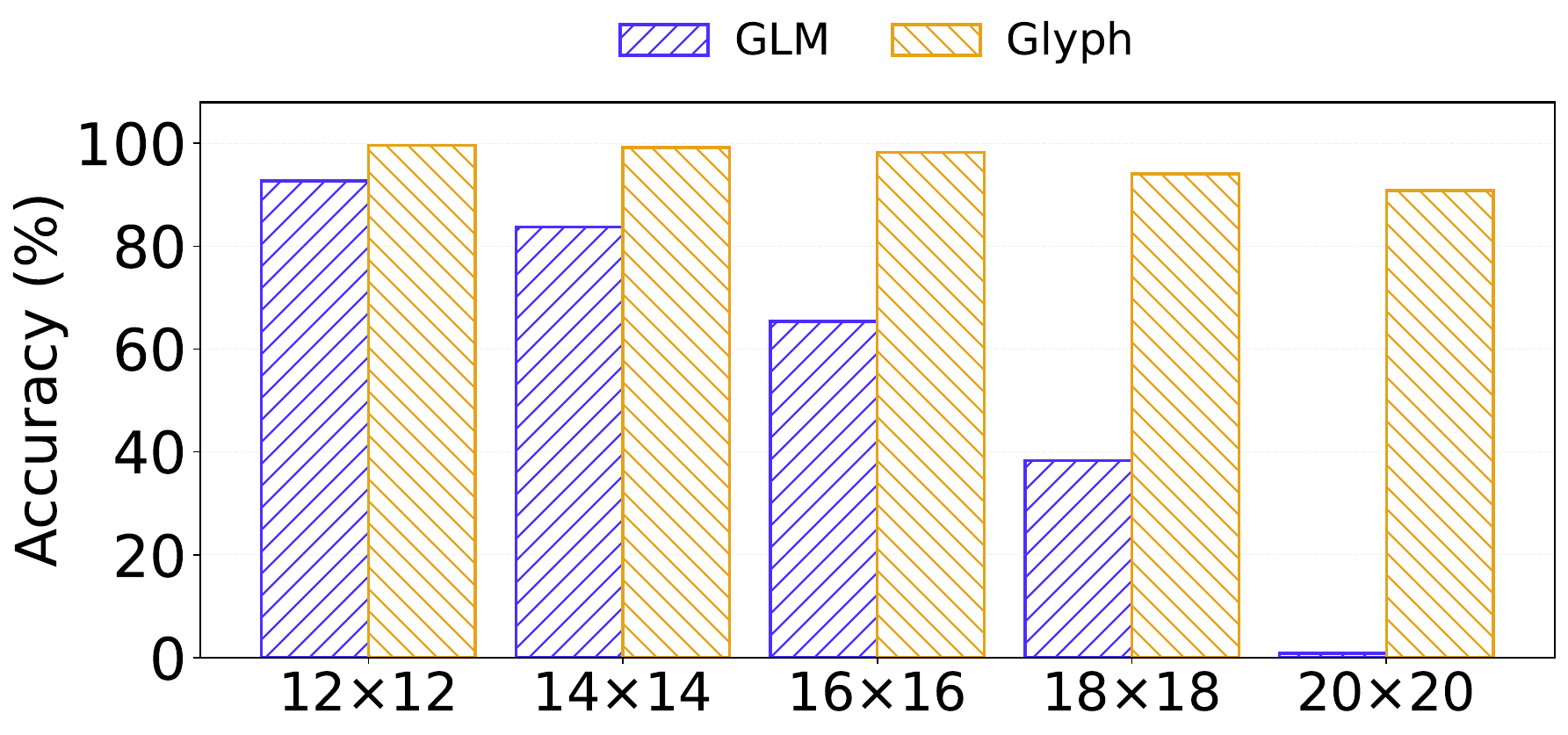}
        \caption{Matched train/eval dimension accuracy}
        \vspace{1.5em}
        \label{fig:transpose_acc_matched}
    \end{subfigure}
    \begin{subfigure}[t]{\linewidth}
        \centering
        \includegraphics[width=\linewidth]{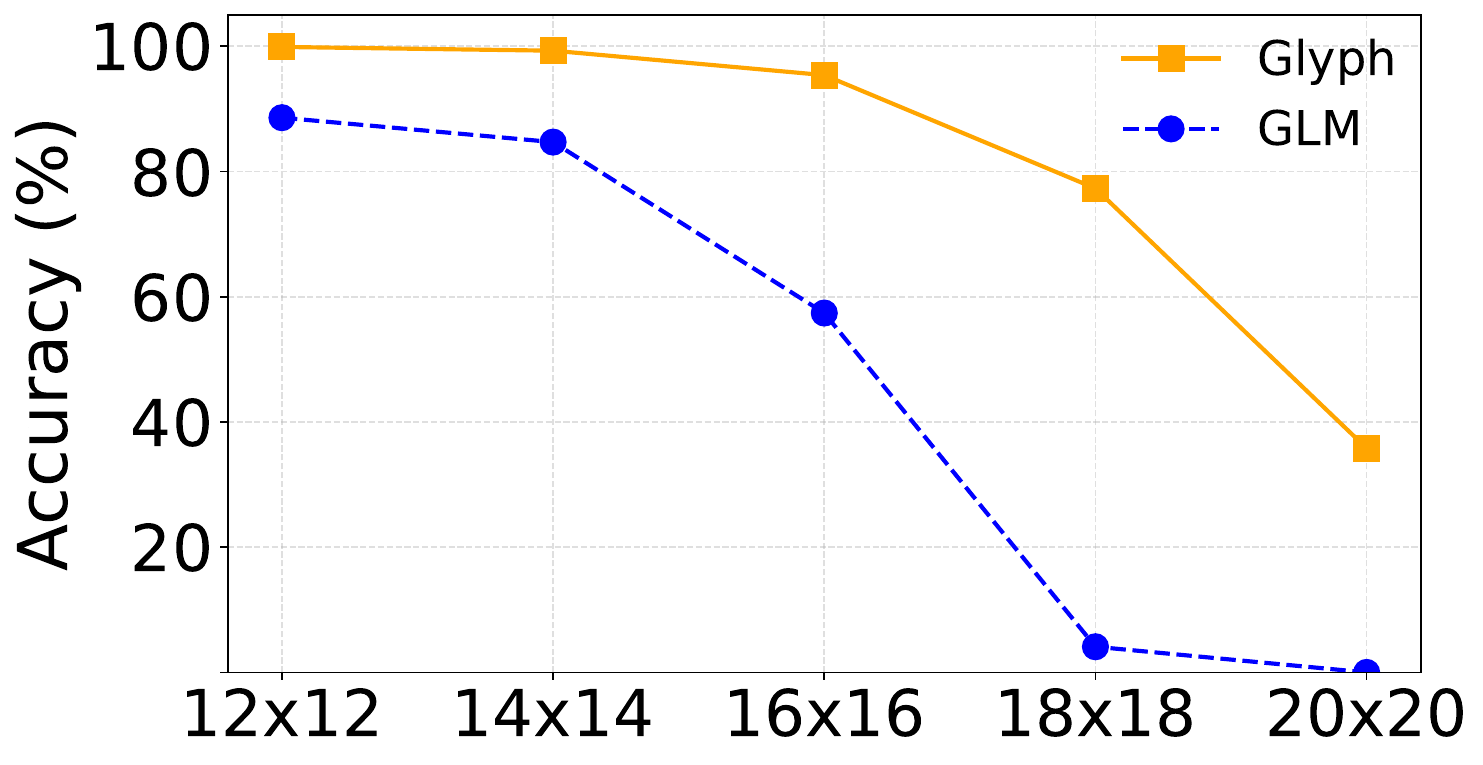}
        \caption{Cross-dimension accuracy}
        \label{fig:transpose_acc_mixed}
    \end{subfigure}
    \caption{Accuracy on the matrix transpose task. 
    (a) Each model is trained and evaluated on the same matrix size. 
    (b) Both models are trained on a mixed dataset of $12\times12$, $14\times14$, 
    and $16\times16$ matrices and evaluated across sizes.}
    \label{fig:transpose_acc}
\end{figure}

\section{Results}

We report results for the three tasks in turn under the comparison setup introduced in Section~\ref{sec:task_suite}. Across this testbed, we examine how the text and visual conditions compare on tasks with explicit 2D structure.

\begin{figure}
    \centering
    \begin{subfigure}[t]{\linewidth}
        \centering
        \includegraphics[width=\linewidth]{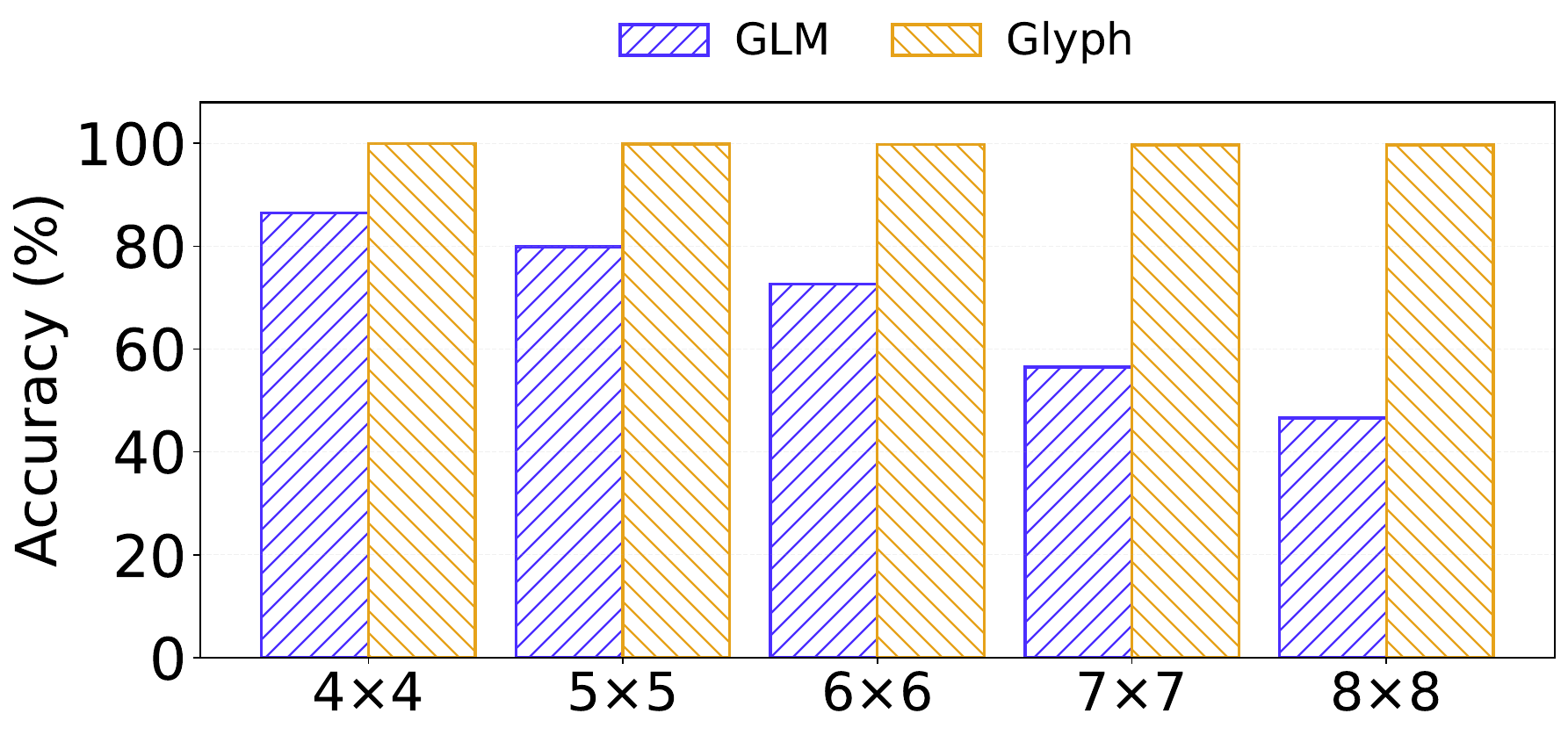}
        \caption{Matched train/eval dimension accuracy}
        \label{fig:conway_acc_matched}
        \vspace{1.5em}
    \end{subfigure}
    \begin{subfigure}[t]{\linewidth}
        \centering
        \includegraphics[width=1.0\linewidth]{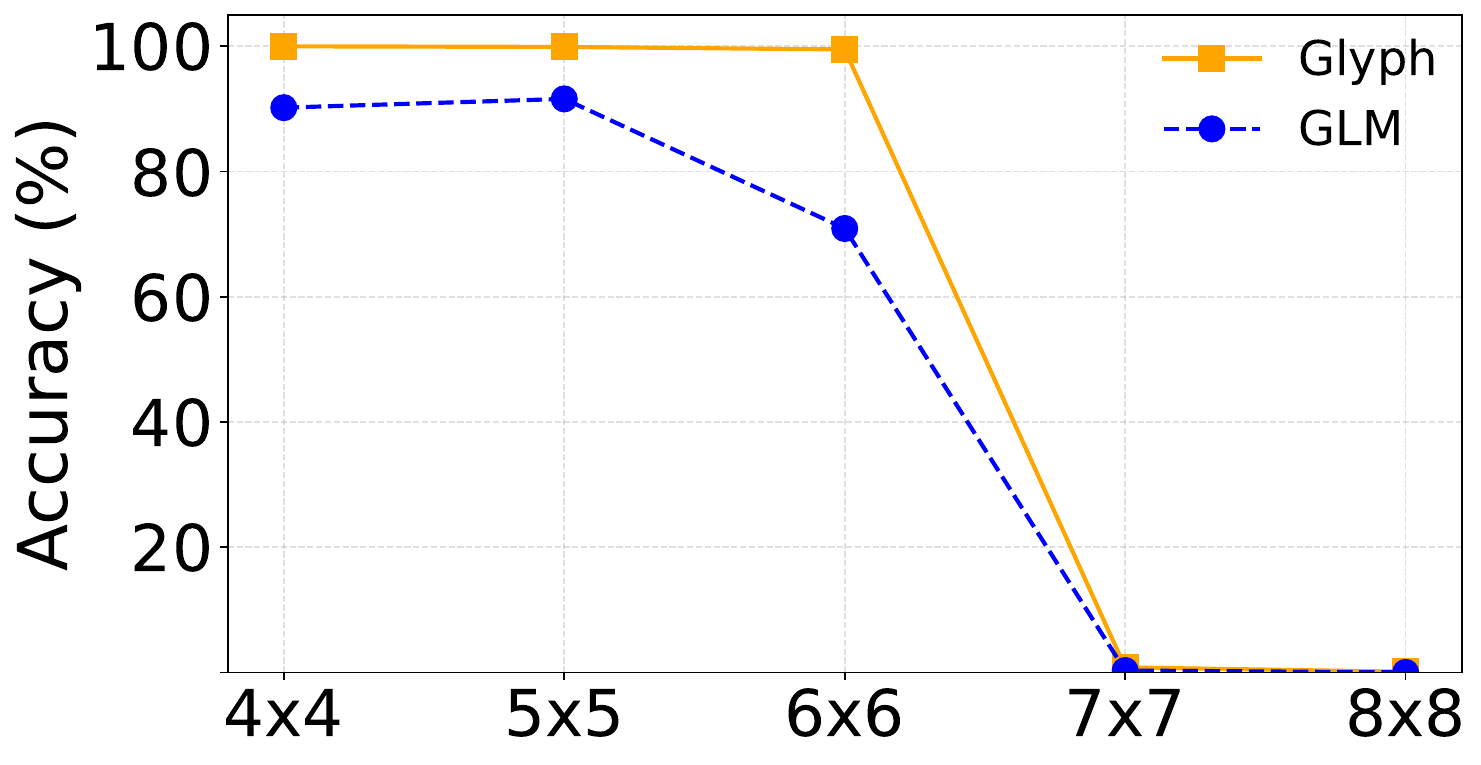}
        \caption{Cross-dimension accuracy}
        \label{fig:conway_acc_mixed}
    \end{subfigure}
    \caption{Accuracy on Conway's Game of Life task. (a) Each model is trained and evaluated on the same grid size. (b) Both models are trained on a mixed dataset of $4\times4$, $5\times5$, and $6\times6$ grids and evaluated across sizes.}
    \label{fig:conway_acc}
\end{figure}

\begin{figure*}[t]
    \centering
    \includegraphics[width=0.98\linewidth]{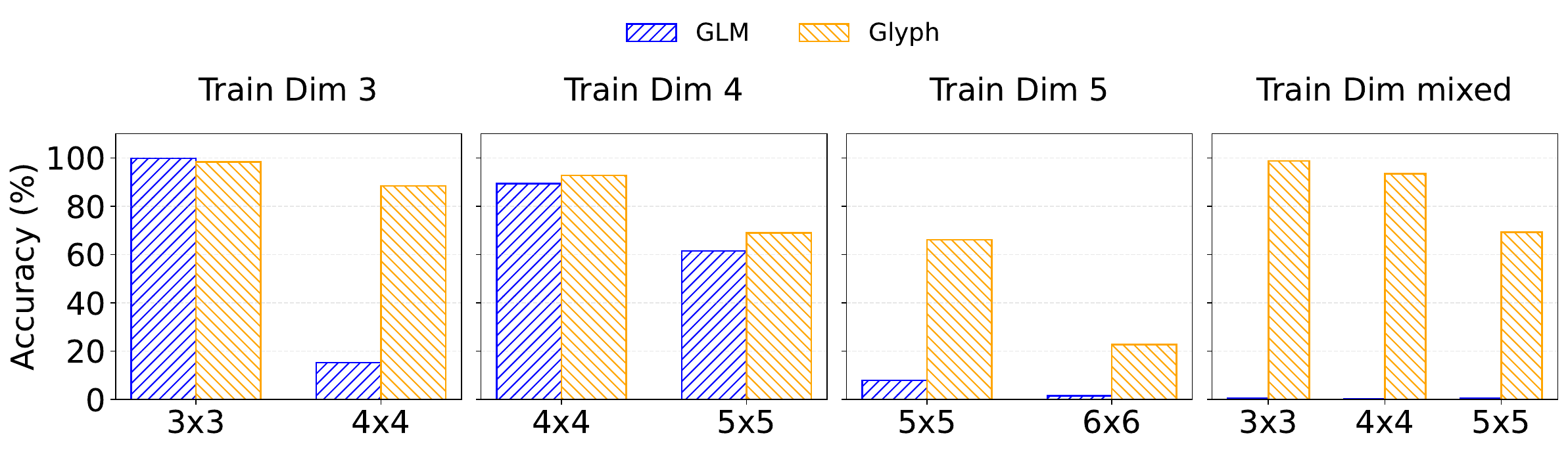}
    \caption{Accuracy on LU decomposition across training and evaluation dimensions. The label above each panel indicates the training setting, with mixed denoting training on a mixture of $3 \times 3$, $4 \times 4$, and $5 \times 5$ matrices. The x-axis shows the evaluation matrix dimension.} 
    \label{fig:lu_acc}
\end{figure*}

\subsection{Matrix Transpose}

Matrix transpose is the most direct task in our testbed, since correctness is determined directly by row--column correspondence in the input layout.

In the matched train/eval setting (Figure~\ref{fig:transpose_acc_matched}), the text and visual conditions already show clearly different trends. Exact-match accuracy under the text condition drops from 92.7\% at $12 \times 12$ to 0.8\% at $20 \times 20$, while the visual condition remains above 90\% across the tested dimensions, decreasing only to 90.8\% at $20 \times 20$. 

The mixed-training setting (Figure~\ref{fig:transpose_acc_mixed}) shows the same pattern more clearly. Within the training range, the visual condition remains near-perfect (99.9\%, 99.8\%, and 95.4\%), whereas exact-match accuracy under the text condition is lower at the same sizes (88.6\%, 74.5\%, and 57.4\%). As evaluation moves beyond the training range, larger unseen sizes become challenging for both conditions. Under the text condition, exact-match accuracy falls to 4.1\% at $18 \times 18$ and 0\% at $20 \times 20$, indicating strong sensitivity to dimension extrapolation in this setting. Although larger unseen sizes remain challenging, this pattern provides preliminary empirical support that presentation choice can affect how the task scales with dimension.

\subsection{Conway's Game of Life}

Conway's Game of Life shifts the structural dependency from row–column correspondence to local neighborhood interaction, and a similar pattern emerges.

In the matched train/eval setting (Figure~\ref{fig:conway_acc_matched}), the text and visual conditions again show clearly different trends: accuracy under the text condition starts at 86.5\% on $4 \times 4$ and falls steadily to 46.6\% at $8 \times 8$, while the visual condition remains above 99.7\% across all tested sizes. This indicates that the difference between the two is also visible in a task defined by local neighborhood interaction.

The mixed-training results (Figure~\ref{fig:conway_acc_mixed}) further refine this picture. Within the training range, the text condition falls from 90.2\% to 70.9\%, while the visual condition remains near-perfect (100.0\% to 99.5\%). Outside the training dimensions, however, the picture is more limited: both conditions lose accuracy as the grid size increases, so the cross-dimension results provide weaker evidence than the within-range comparison. The pattern observed in matrix transpose therefore carries over most clearly within the training range, while extrapolation to larger Conway grids remains challenging for both presentations.

\subsection{LU Decomposition}

In LU decomposition, the role of structure differs from the previous two tasks. Correctness is not determined by a single spatial readout, but by a sequence of dependent row--column operations. The task therefore tests a setting in which 2D structure remains relevant throughout a multi-step transformation, rather than only at the input-output mapping.

Figure~\ref{fig:lu_acc} shows that the separation between the two conditions varies across training settings and evaluation dimensions. In the single-dimension training settings, the two are often comparable at or near the training dimension, while the gap becomes more visible once evaluation moves beyond it. When trained on $3 \times 3$ matrices, both are near-perfect at $3 \times 3$ (99.8\% under text and 98.3\% under visual), but at $4 \times 4$ accuracy drops to 15.3\% under text while remaining at 88.4\% under visual. A related pattern appears when training on $4 \times 4$: both remain strong at the training size (89.4\% under text and 92.8\% under visual), while at $5 \times 5$ accuracy falls to 61.5\% and 69.0\%, respectively.

The mixed-training setting is where the dependence on presentation format is most apparent. Under the text condition, accuracy remains near zero across all training dimensions (0.6\%, 0.3\%, and 0.4\%), while visual condition maintains substantial accuracy at the same sizes (98.8\%, 93.5\%, and 69.2\%). Taken together, these results suggest that even for a task requiring sustained 2D structure across dependent steps, the presentation choice remains consequential.

\section{Supplementary Experiments}

Section~\ref{sec:within_probe} uses a within-visual probe to examine whether the observed difference is explained by image encoding alone. Section~\ref{app:qwen_comparison} uses a supplementary Qwen experiment to test whether the question of input presentation remains relevant beyond the GLM/Glyph setup. The final subsection analyzes task-structured error patterns and relates them to each task's underlying 2D structure.

\subsection{A within-visual probe for interpretation}
\label{sec:within_probe}

\begin{figure}[t]
    \centering
    \includegraphics[width=1.0\linewidth]{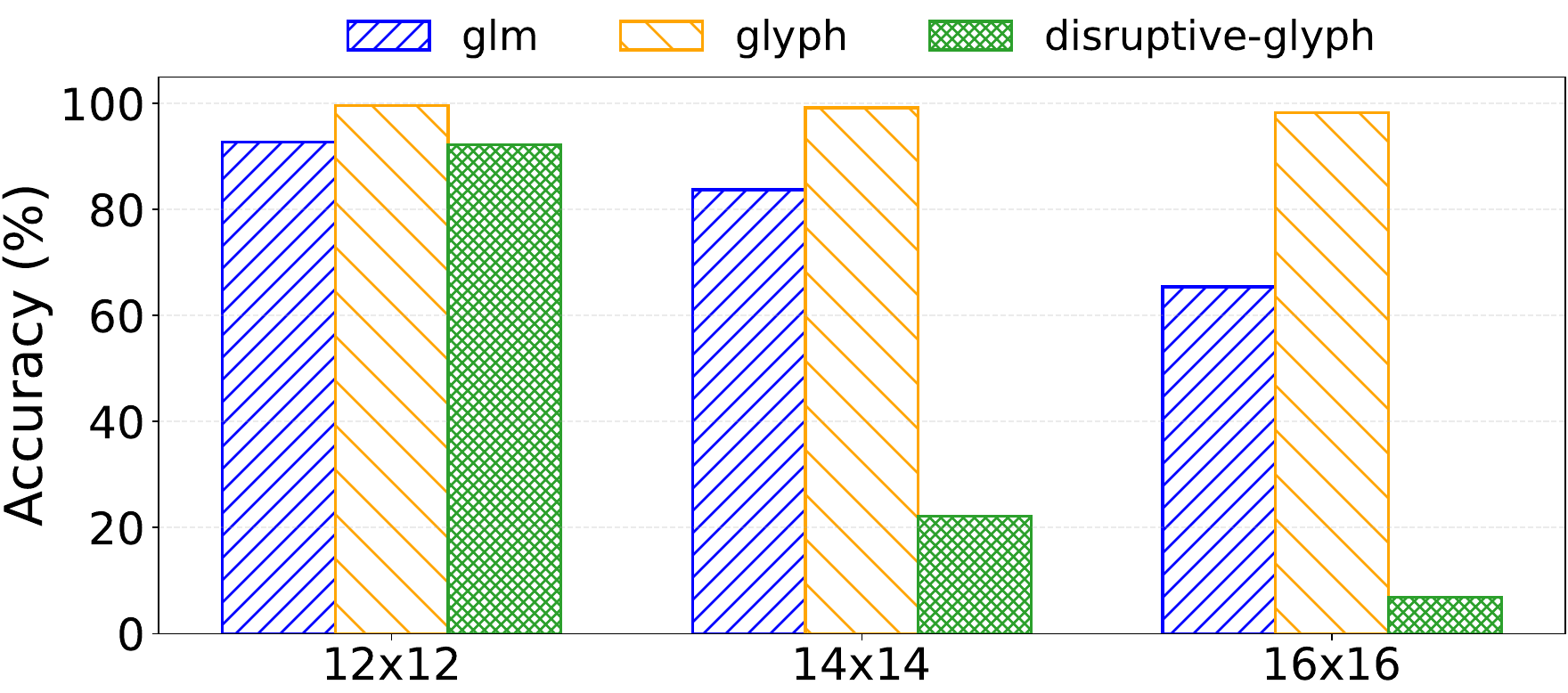}
    \caption{Accuracy on matrix transpose under the within-visual probe, where each model is trained and evaluated at the same matrix size.}
    \label{fig:within_visual_probe}
    \vspace{-1pt}
\end{figure}

The preceding experiments do not fully separate the effect of preserving 2D layout from the additional capabilities introduced by the vision-language pathway. To better separate these factors, we add a smaller probe on matrix transpose in which only the visual rendering is changed. Specifically, we compare a native 2D matrix rendering with a layout-disrupted rendering that presents the same row-major textual content as an image. We use transpose because it provides the most direct test of row--column organization in our task suite. Details of the two rendering strategies are provided in Appendix~\ref{app:rendering_details}.

Figure~\ref{fig:within_visual_probe} shows a clear separation among the three conditions: across all tested dimensions, the native 2D rendering performs best, the serialized text condition remains in the middle, and the layout-disrupted rendering performs worst. Although limited to a single task, this probe provides a more constrained comparison for interpreting the main result: within the same visual pathway, changing the input layout substantially affects transpose performance. This does not isolate a unique causal mechanism, but it suggests that, in this setting, representation choice matters beyond the use of visual encoding alone.

\subsection{Supplementary Evidence under the Mixed-Training Transpose Setting}
\label{app:qwen_comparison}

To examine whether the difference between 1D serialization and native 2D layout is limited to the specific model setup used in the main experiments, we use the mixed-training transpose setting as a supplementary probe with Qwen models. In this setting, the text condition uses Qwen2.5-3B base~\citep{yang2024qwen2} and receives the matrix as a 1D serialized text input, while the visual condition uses Qwen2.5-VL-3B-Instruct~\citep{bai2025qwen2} and receives the same underlying matrix rendered in its native 2D layout. Both conditions are trained on a mixture of $4\times4$, $6\times6$, and $8\times8$ transpose examples and evaluated separately on each size. Experiment details are provided in Appendix~\ref{app:qwen_transpose}.

As shown in Figure~\ref{fig:qwen_comparison}, a similar pattern appears. The visual condition sustains high exact-match accuracy as matrix size increases, achieving 100.0\%, 99.8\%, and 97.6\% accuracy at $4\times4$, $6\times6$, and $8\times8$, respectively. In contrast, the text condition drops sharply as matrix size increases, from 97.3\% at $4\times4$ to 34.0\% at $6\times6$ and 0\% at $8\times8$. This supplementary result suggests that the question of input presentation choice remains relevant beyond the original experimental setup.

\begin{figure}[t]
    \centering
    \includegraphics[width=\linewidth]{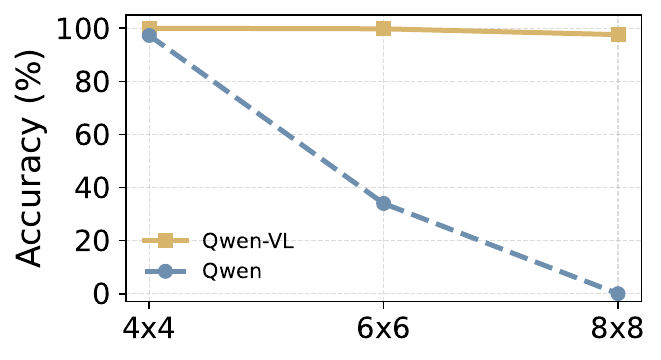}
    \caption{Supplementary results on matrix transpose under the mixed-training setting.}
    \label{fig:qwen_comparison}
\end{figure}

\subsection{Task-structured error patterns}

Aggregate accuracy shows that 1D serialization and native 2D layout behave differently, but it does not by itself explain how those differences arise within each task. We therefore examine finer-grained error patterns. Across the three tasks, the gap does not appear as unstructured noise; instead, it follows the particular kind of 2D structure each task depends on.

\begin{figure*}[t]
    \centering
    \includegraphics[width=0.99\linewidth]{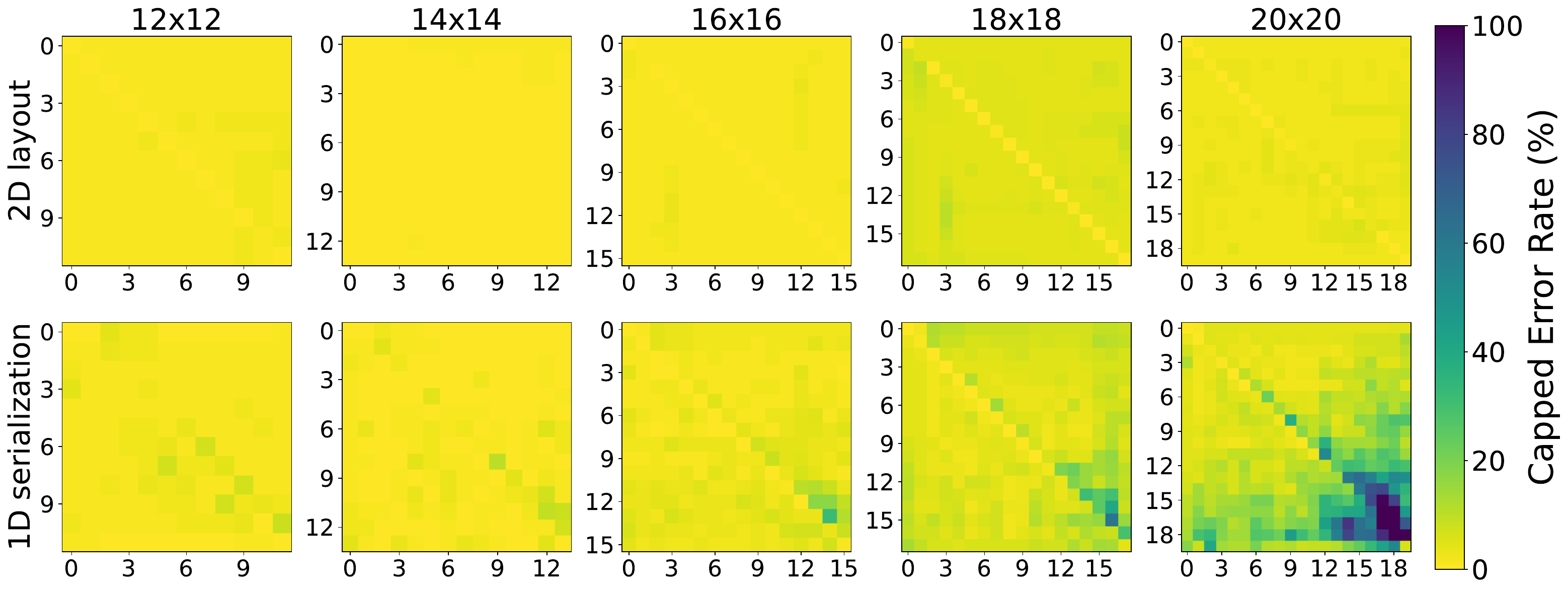}
    \caption{Cell-level transpose error heatmaps across matrix sizes for 2D layout (top) and 1D serialization (bottom). The two representations exhibit visibly different spatial error patterns as matrix size increases.}
    \label{fig:transpose_err_heatmap}
\end{figure*}

Figure~\ref{fig:transpose_err_heatmap} shows the spatial concentration of errors in matrix transpose. Under 1D serialization, errors become increasingly concentrated in the lower-right region of the matrix as dimension grows, whereas under 2D layout they remain uniformly low across sizes. This concentration grows more pronounced as matrix size increases, with the lower-right region becoming increasingly unreliable under 1D serialization.

\begin{figure}[t]
    \centering
    \includegraphics[width=1\linewidth]{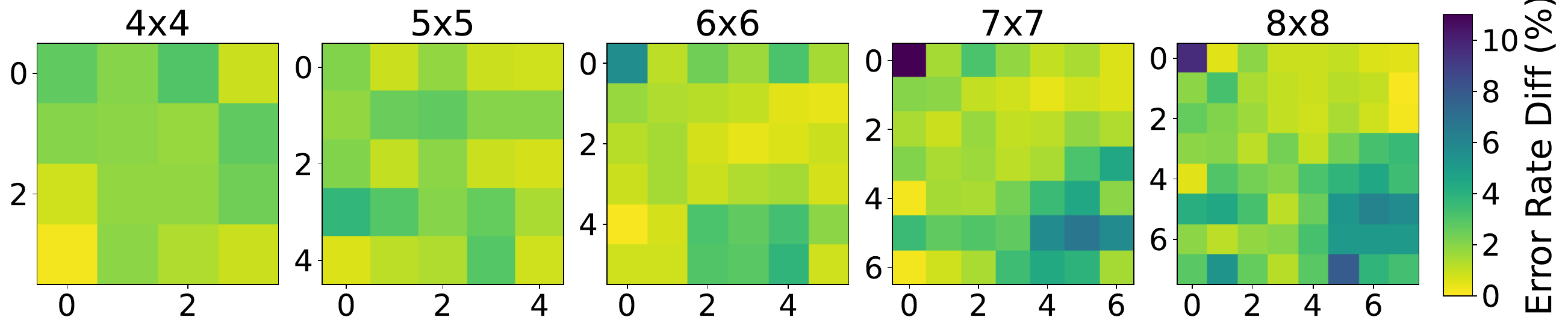}
    \caption{Cell-wise error-rate difference heatmaps for Conway's Game of Life across grid sizes, computed as the GLM error rate minus the Glyph error rate. Positive values indicate regions where 1D serialization incurs higher error than the native 2D layout.}
    \label{fig:conway_err_diff_heatmap}
    \vspace{-1pt}
\end{figure}

A different pattern appears in Conway's Game of Life, shown in Figure~\ref{fig:conway_err_diff_heatmap}. Here the error-rate difference remains relatively diffuse rather than concentrating in a single region, even as its magnitude grows with board size. This is consistent with the nature of the task: since every cell depends on its local neighborhood, the cost of losing spatial adjacency under 1D serialization is not localized to a particular region but distributed across the grid.

\begin{figure*}[t]
    \centering
    \includegraphics[width=\linewidth]{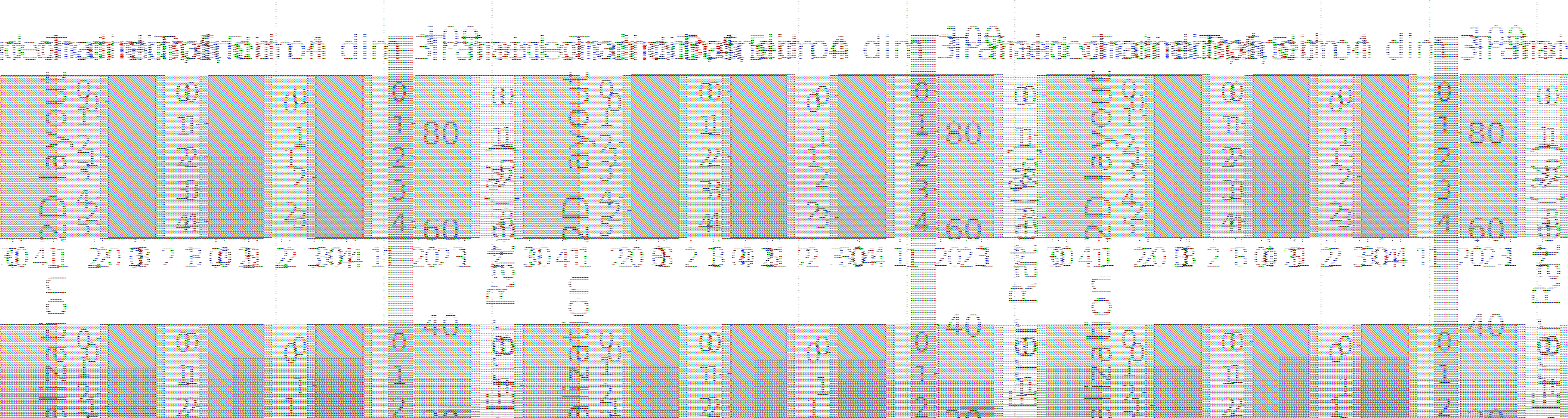}
    \caption{Cell-level error heatmaps for LU decomposition across training configurations for 2D layout (top) and 1D serialization (bottom). Darker cells indicate higher error rates at the corresponding matrix positions.}
    \label{fig:err_heatmap}
\end{figure*}

LU decomposition shows a different error pattern, as illustrated in Figure~\ref{fig:err_heatmap}. Under 1D serialization, errors are more concentrated in later rows and columns, especially on larger evaluation sizes, whereas under 2D layout they remain milder and less sharply localized. At a descriptive level, this matches the structure of the task, where correctness depends on a sequence of dependent elimination steps. A qualitative example is shown in Appendix~\ref{app:lu_case_study}: under 2D layout, the model's reasoning is more closely organized around the visible matrix structure, while under 1D serialization it relies more on symbolic index tracking.

Taken together, these results suggest that the error patterns under 1D serialization differ across tasks. These differences are not captured by aggregate accuracy alone; instead, the observed errors are consistent with the kind of 2D organization emphasized by each task, such as global row--column alignment in transpose, local spatial adjacency in Conway's Game of Life, and multi-step row--column transformations in LU decomposition.

\section{Conclusion}



This paper examines what happens to problems originally defined by explicit 2D structure when they are instead presented through 1D serialization. Across the controlled synthetic testbed studied here, 1D serialization is consistently associated with sharper accuracy loss and more spatially structured error patterns across most settings. Supplementary analyses, including a within-visual probe and an additional presentation setting, provide further context for interpreting this pattern. For the layout-defined tasks studied here, our results suggest that reducing inputs to ordinary 1D serializations should not be treated as a representation-neutral default, but rather as a design choice whose consequences warrant deliberate consideration.


\section{Limitations}

This is a system-level representation comparison rather than a fully isolated test of layout alone. The textual and visual conditions still differ in modality, visual front end, pretraining-derived priors, tokenization regime, and input representation, so the observed pattern should not be interpreted as evidence for a unique causal mechanism. The within-visual probe in Section~\ref{sec:within_probe} helps narrow this interpretation, but remains supportive rather than definitive evidence.

The testbed is intentionally narrow, covering three synthetic tasks chosen because their relevant relations are organized by explicit 2D structure. This focus limits coverage across task types, and leaves open how far the observed pattern extends to broader classes of structured problems, real-world inputs, and other model settings. Broader task coverage and tighter controls separating layout, modality, tokenization, and encoder effects are needed to assess the generality of the findings.

\clearpage
\bibliography{custom}

@article{vaswani2017attention,
  title        = {Attention Is All You Need},
  author       = {Vaswani, Ashish and Shazeer, Noam and Parmar, Niki and Uszkoreit, Jakob and Jones, Llion and Gomez, Aidan N. and Kaiser, Lukasz and Polosukhin, Illia},
  journal      = {CoRR},
  volume       = {abs/1706.03762},
  year         = {2017},
  url          = {https://doi.org/10.48550/arXiv.1706.03762},
  doi          = {10.48550/arXiv.1706.03762},
  eprinttype   = {arXiv},
  eprint       = {1706.03762}
}

@article{brown2020gpt3,
  title        = {Language Models are Few-Shot Learners},
  author       = {Brown, Tom B. and Mann, Benjamin and Ryder, Nick and Subbiah, Melanie and Kaplan, Jared and Dhariwal, Prafulla and Neelakantan, Arvind and Shyam, Pranav and Sastry, Girish and Askell, Amanda and Agarwal, Sandhini and Herbert-Voss, Ariel and Krueger, Gretchen and Henighan, Tom and Child, Rewon and Ramesh, Aditya and Ziegler, Daniel M. and Wu, Jeffrey and Winter, Clemens and Hesse, Christopher and Chen, Mark and Sigler, Eric and Litwin, Mateusz and Gray, Scott and Chess, Benjamin and Clark, Jack and Berner, Christopher and McCandlish, Sam and Radford, Alec and Sutskever, Ilya and Amodei, Dario},
  journal      = {CoRR},
  volume       = {abs/2005.14165},
  year         = {2020},
  url          = {https://doi.org/10.48550/arXiv.2005.14165},
  doi          = {10.48550/arXiv.2005.14165},
  eprinttype   = {arXiv},
  eprint       = {2005.14165}
}

@article{gardner1970fantastic,
  title={The Fantastic Combinations of John Conway's New Solitaire Game 'Life'.},
  author={Gardner, M.},
  journal={Scientific American},
  volume={223},
  number={4},
  pages={120--123},
  year={1970}
}

@article{lee2022pix2struct,
  title        = {Pix2Struct: Screenshot Parsing as Pretraining for Visual Language Understanding},
  author       = {Lee, Kenton and Joshi, Mandar and Turc, Iulia and Hu, Hexiang and Liu, Fangyu and Eisenschlos, Julian and Khandelwal, Urvashi and Shaw, Peter and Chang, Ming-Wei and Toutanova, Kristina},
  journal      = {CoRR},
  volume       = {abs/2210.03347},
  year         = {2022},
  url          = {https://doi.org/10.48550/arXiv.2210.03347},
  doi          = {10.48550/arXiv.2210.03347},
  eprinttype   = {arXiv},
  eprint       = {2210.03347}
}

@article{huang2022layoutlmv3,
  title        = {LayoutLMv3: Pre-training for Document AI with Unified Text and Image Masking},
  author       = {Huang, Yupan and Lv, Tengchao and Cui, Lei and Lu, Yutong and Wei, Furu},
  journal      = {CoRR},
  volume       = {abs/2204.08387},
  year         = {2022},
  url          = {https://doi.org/10.48550/arXiv.2204.08387},
  doi          = {10.48550/arXiv.2204.08387},
  eprinttype   = {arXiv},
  eprint       = {2204.08387}
}

@article{herzig2020tapas,
  title        = {TAPAS: Weakly Supervised Table Parsing via Pre-training},
  author       = {Herzig, Jonathan and Nowak, Pawel Krzysztof and M{\"u}ller, Thomas and Piccinno, Francesco and Eisenschlos, Julian Martin},
  journal      = {CoRR},
  volume       = {abs/2004.02349},
  year         = {2020},
  url          = {https://doi.org/10.48550/arXiv.2004.02349},
  doi          = {10.48550/arXiv.2004.02349},
  eprinttype   = {arXiv},
  eprint       = {2004.02349}
}

@article{kim2021donut,
  title        = {OCR-free Document Understanding Transformer},
  author       = {Kim, Geewook and Hong, Teakgyu and Yim, Moonbin and Nam, Jeongyeon and Park, Jinyoung and Yim, Jinyeong and Hwang, Wonseok and Yun, Sangdoo and Han, Dongyoon and Park, Seunghyun},
  journal      = {CoRR},
  volume       = {abs/2111.15664},
  year         = {2021},
  url          = {https://doi.org/10.48550/arXiv.2111.15664},
  doi          = {10.48550/arXiv.2111.15664},
  eprinttype   = {arXiv},
  eprint       = {2111.15664}
}

@article{cheng2025glyph,
  title        = {Glyph: Scaling Context Windows via Visual-Text Compression},
  author       = {Cheng, Jiale and Liu, Yusen and Zhang, Xinyu and Fei, Yulin and Hong, Wenyi and Lyu, Ruiliang and Wang, Weihan and Su, Zhe and Gu, Xiaotao and Liu, Xiao and Bai, Yushi and Tang, Jie and Wang, Hongning and Huang, Minlie},
  journal      = {CoRR},
  volume       = {abs/2510.17800},
  year         = {2025},
  url          = {https://doi.org/10.48550/arXiv.2510.17800},
  doi          = {10.48550/arXiv.2510.17800},
  eprinttype   = {arXiv},
  eprint       = {2510.17800}
}

@article{wei2025deepseekocr,
  title        = {DeepSeek-OCR: Contexts Optical Compression},
  author       = {Wei, Haoran and Sun, Yaofeng and Li, Yukun},
  journal      = {CoRR},
  volume       = {abs/2510.18234},
  year         = {2025},
  url          = {https://doi.org/10.48550/arXiv.2510.18234},
  doi          = {10.48550/arXiv.2510.18234},
  eprinttype   = {arXiv},
  eprint       = {2510.18234}
}

@inproceedings{hu2021lora,
  author    = {Edward J. Hu and Yelong Shen and Phillip Wallis and Zeyuan Allen{-}Zhu and Yuanzhi Li and Shean Wang and Lu Wang and Weizhu Chen},
  title     = {LoRA: Low-Rank Adaptation of Large Language Models},
  booktitle = {The Tenth International Conference on Learning Representations, {ICLR} 2022, Virtual Event, April 25-29, 2022},
  publisher = {OpenReview.net},
  year      = {2022},
  url       = {https://openreview.net/forum?id=nZeVKeeFYf9},
  bibsource = {dblp computer science bibliography, https://dblp.org}
}

@article{yu2025dapo,
  title={Dapo: An open-source llm reinforcement learning system at scale},
  author={Yu, Qiying and Zhang, Zheng and Zhu, Ruofei and Yuan, Yufeng and Zuo, Xiaochen and Yue, Yu and Fan, Tiantian and Liu, Gaohong and Liu, Lingjun and Liu, Xin and others},
  journal={arXiv preprint arXiv:2503.14476},
  year={2025}
}

@article{glm2024chatglm,
  title   = {ChatGLM: A Family of Large Language Models from GLM-130B to GLM-4 All Tools},
  author  = {Team GLM and : and Aohan Zeng and Bin Xu and Bowen Wang and Chenhui Zhang and Da Yin and Dan Zhang and Diego Rojas and Guanyu Feng and Hanlin Zhao and Hanyu Lai and Hao Yu and Hongning Wang and Jiadai Sun and Jiajie Zhang and Jiale Cheng and Jiayi Gui and Jie Tang and Jing Zhang and Jingyu Sun and Juanzi Li and Lei Zhao and Lindong Wu and Lucen Zhong and Mingdao Liu and Minlie Huang and Peng Zhang and Qinkai Zheng and Rui Lu and Shuaiqi Duan and Shudan Zhang and Shulin Cao and Shuxun Yang and Weng Lam Tam and Wenyi Zhao and Xiao Liu and Xiao Xia and Xiaohan Zhang and Xiaotao Gu and Xin Lv and Xinghan Liu and Xinyi Liu and Xinyue Yang and Xixuan Song and Xunkai Zhang and Yifan An and Yifan Xu and Yilin Niu and Yuantao Yang and Yueyan Li and Yushi Bai and Yuxiao Dong and Zehan Qi and Zhaoyu Wang and Zhen Yang and Zhengxiao Du and Zhenyu Hou and Zihan Wang},
  year    = {2024},
  journal = {arXiv preprint arXiv: 2406.12793}
}

@article{shao2024deepseekmath,
  title={Deepseekmath: Pushing the limits of mathematical reasoning in open language models},
  author={Shao, Zhihong and Wang, Peiyi and Zhu, Qihao and Xu, Runxin and Song, Junxiao and Bi, Xiao and Zhang, Haowei and Zhang, Mingchuan and Li, YK and Wu, Yang and others},
  journal={arXiv preprint arXiv:2402.03300},
  url= {https://doi.org/10.48550/arXiv.2402.03300},
  year={2024}
}

@inproceedings{sui2024table,
  title={Table meets llm: Can large language models understand structured table data? a benchmark and empirical study},
  author={Sui, Yuan and Zhou, Mengyu and Zhou, Mingjie and Han, Shi and Zhang, Dongmei},
  booktitle={Proceedings of the 17th ACM International Conference on Web Search and Data Mining},
  pages={645--654},
  year={2024}
}

@inproceedings{wang2021tuta,
  title={Tuta: Tree-based transformers for generally structured table pre-training},
  author={Wang, Zhiruo and Dong, Haoyu and Jia, Ran and Li, Jia and Fu, Zhiyi and Han, Shi and Zhang, Dongmei},
  booktitle={Proceedings of the 27th ACM SIGKDD Conference on Knowledge Discovery \& Data Mining},
  pages={1780--1790},
  year={2021}
}

@inproceedings{yin2020tabert,
  title={TaBERT: Pretraining for joint understanding of textual and tabular data},
  author={Yin, Pengcheng and Neubig, Graham and Yih, Wen-tau and Riedel, Sebastian},
  booktitle={Proceedings of the 58th annual meeting of the association for computational linguistics},
  pages={8413--8426},
  year={2020}
}

@article{zhao2025vtcbench,
  title={VTCBench: Can Vision-Language Models Understand Long Context with Vision-Text Compression?},
  author={Zhao, Hongbo and Wang, Meng and Zhu, Fei and Liu, Wenzhuo and Ni, Bolin and Zeng, Fanhu and Meng, Gaofeng and Zhang, Zhaoxiang},
  journal={arXiv preprint arXiv:2512.15649},
  url= {https://doi.org/10.48550/arXiv.2512.15649},
  year={2025}
}

@article{fatemi2023talk,
  title={Talk like a graph: Encoding graphs for large language models},
  author={Fatemi, Bahare and Halcrow, Jonathan and Perozzi, Bryan},
  journal={arXiv preprint arXiv:2310.04560},
  url= {https://doi.org/10.48550/arXiv.2310.04560},
  year={2023}
}

@inproceedings{yin-etal-2025-talk,
    title = "How to Talk to Language Models: Serialization Strategies for Structured Entity Matching",
    author = "Yin, Haoteng  and
      Kim, Jinha  and
      Mathur, Prashant  and
      Sarker, Krishanu  and
      Bansal, Vidit",
    editor = "Chiruzzo, Luis  and
      Ritter, Alan  and
      Wang, Lu",
    booktitle = "Findings of the Association for Computational Linguistics: NAACL 2025",
    month = apr,
    year = "2025",
    address = "Albuquerque, New Mexico",
    publisher = "Association for Computational Linguistics",
    url = "https://aclanthology.org/2025.findings-naacl.437/",
    doi = "10.18653/v1/2025.findings-naacl.437",
    pages = "7851--7865",
    ISBN = "979-8-89176-195-7",
}

@article{izadi2025visualstructures,
  title   = {Visual Structures Helps Visual Reasoning: Addressing the Binding Problem in VLMs},
  author  = {Amirmohammad Izadi and Mohammad Ali Banayeeanzade and Fatemeh Askari and Ali Rahimiakbar and Mohammad Mahdi Vahedi and Hosein Hasani and Mahdieh Soleymani Baghshah},
  year    = {2025},
  journal = {arXiv preprint arXiv: 2506.22146}
}

@article{wei2024gita,
  title   = {GITA: Graph to Visual and Textual Integration for Vision-Language Graph Reasoning},
  author  = {Yanbin Wei and Shuai Fu and Weisen Jiang and Zejian Zhang and Zhixiong Zeng and Qi Wu and James T. Kwok and Yu Zhang},
  year    = {2024},
  journal = {arXiv preprint arXiv: 2402.02130}
}

@article{dantart2026topo0rag0,
  title   = {Topo-RAG: Topology-aware retrieval for hybrid text-table documents},
  author  = {Alex Dantart and Marco Kóvacs-Navarro},
  year    = {2026},
  journal = {arXiv preprint arXiv: 2601.10215}
}

@article{liu2026vista0bench0,
  title   = {VISTA-Bench: Do Vision-Language Models Really Understand Visualized Text as Well as Pure Text?},
  author  = {Qing'an Liu and Juntong Feng and Yuhao Wang and Xinzhe Han and Yujie Cheng and Yue Zhu and Haiwen Diao and Yunzhi Zhuge and Huchuan Lu},
  year    = {2026},
  journal = {arXiv preprint arXiv: 2602.04802}
}

@article{sun2026reading0,
  title   = {Reading, Not Thinking: Understanding and Bridging the Modality Gap When Text Becomes Pixels in Multimodal LLMs},
  author  = {Kaiser Sun and Xiaochuang Yuan and Hongjun Liu and Chen Zhao and Cheng Zhang and Mark Dredze and Fan Bai},
  year    = {2026},
  journal = {arXiv preprint arXiv: 2603.09095}
}

@article{deepseek-ai2025deepseek0r10,
  title   = {DeepSeek-R1: Incentivizing Reasoning Capability in LLMs via Reinforcement Learning},
  author  = {DeepSeek-AI and Daya Guo and Dejian Yang and Haowei Zhang and Junxiao Song and Peiyi Wang and Qihao Zhu and Runxin Xu and Ruoyu Zhang and Shirong Ma and Xiao Bi and Xiaokang Zhang and Xingkai Yu and Yu Wu and Z. F. Wu and Zhibin Gou and Zhihong Shao and Zhuoshu Li and Ziyi Gao and Aixin Liu and Bing Xue and Bingxuan Wang and Bochao Wu and Bei Feng and Chengda Lu and Chenggang Zhao and Chengqi Deng and Chenyu Zhang and Chong Ruan and Damai Dai and Deli Chen and Dongjie Ji and Erhang Li and Fangyun Lin and Fucong Dai and Fuli Luo and Guangbo Hao and Guanting Chen and Guowei Li and H. Zhang and Han Bao and Hanwei Xu and Haocheng Wang and Honghui Ding and Huajian Xin and Huazuo Gao and Hui Qu and Hui Li and Jianzhong Guo and Jiashi Li and Jiawei Wang and Jingchang Chen and Jingyang Yuan and Junjie Qiu and Junlong Li and J. L. Cai and Jiaqi Ni and Jian Liang and Jin Chen and Kai Dong and Kai Hu and Kaige Gao and Kang Guan and Kexin Huang and Kuai Yu and Lean Wang and Lecong Zhang and Liang Zhao and Litong Wang and Liyue Zhang and Lei Xu and Leyi Xia and Mingchuan Zhang and Minghua Zhang and Minghui Tang and Meng Li and Miaojun Wang and Mingming Li and Ning Tian and Panpan Huang and Peng Zhang and Qiancheng Wang and Qinyu Chen and Qiushi Du and Ruiqi Ge and Ruisong Zhang and Ruizhe Pan and Runji Wang and R. J. Chen and R. L. Jin and Ruyi Chen and Shanghao Lu and Shangyan Zhou and Shanhuang Chen and Shengfeng Ye and Shiyu Wang and Shuiping Yu and Shunfeng Zhou and Shuting Pan and S. S. Li and Shuang Zhou and Shaoqing Wu and Shengfeng Ye and Tao Yun and Tian Pei and Tianyu Sun and T. Wang and Wangding Zeng and Wanjia Zhao and Wen Liu and Wenfeng Liang and Wenjun Gao and Wenqin Yu and Wentao Zhang and W. L. Xiao and Wei An and Xiaodong Liu and Xiaohan Wang and Xiaokang Chen and Xiaotao Nie and Xin Cheng and Xin Liu and Xin Xie and Xingchao Liu and Xinyu Yang and Xinyuan Li and Xuecheng Su and Xuheng Lin and X. Q. Li and Xiangyue Jin and Xiaojin Shen and Xiaosha Chen and Xiaowen Sun and Xiaoxiang Wang and Xinnan Song and Xinyi Zhou and Xianzu Wang and Xinxia Shan and Y. K. Li and Y. Q. Wang and Y. X. Wei and Yang Zhang and Yanhong Xu and Yao Li and Yao Zhao and Yaofeng Sun and Yaohui Wang and Yi Yu and Yichao Zhang and Yifan Shi and Yiliang Xiong and Ying He and Yishi Piao and Yisong Wang and Yixuan Tan and Yiyang Ma and Yiyuan Liu and Yongqiang Guo and Yuan Ou and Yuduan Wang and Yue Gong and Yuheng Zou and Yujia He and Yunfan Xiong and Yuxiang Luo and Yuxiang You and Yuxuan Liu and Yuyang Zhou and Y. X. Zhu and Yanhong Xu and Yanping Huang and Yaohui Li and Yi Zheng and Yuchen Zhu and Yunxian Ma and Ying Tang and Yukun Zha and Yuting Yan and Z. Z. Ren and Zehui Ren and Zhangli Sha and Zhe Fu and Zhean Xu and Zhenda Xie and Zhengyan Zhang and Zhewen Hao and Zhicheng Ma and Zhigang Yan and Zhiyu Wu and Zihui Gu and Zijia Zhu and Zijun Liu and Zilin Li and Ziwei Xie and Ziyang Song and Zizheng Pan and Zhen Huang and Zhipeng Xu and Zhongyu Zhang and Zhen Zhang},
  year    = {2025},
  journal = {arXiv preprint arXiv: 2501.12948}
}

@article{yang2025depth0breadth,
  title   = {Depth-Breadth Synergy in RLVR: Unlocking LLM Reasoning Gains with Adaptive Exploration},
  author  = {Zhicheng Yang and Zhijiang Guo and Yinya Huang and Yongxin Wang and Dongchun Xie and Hanhui Li and Yiwei Wang and Xiaodan Liang and Jing Tang},
  year    = {2025},
  journal = {arXiv preprint arXiv: 2508.13755}
}

@inproceedings{sheng2025hybridflow,
  title={Hybridflow: A flexible and efficient rlhf framework},
  author={Sheng, Guangming and Zhang, Chi and Ye, Zilingfeng and Wu, Xibin and Zhang, Wang and Zhang, Ru and Peng, Yanghua and Lin, Haibin and Wu, Chuan},
  booktitle={Proceedings of the Twentieth European Conference on Computer Systems},
  pages={1279--1297},
  year={2025}
}

@article{yang2024qwen2,
  title={Qwen2.5 Technical Report},
  author={Yang, An and Yang, Baosong and Zhang, Beichen and Hui, Binyuan and Zheng, Bo and Yu, Bowen and Li, Chengyuan and Liu, Dayiheng and Huang, Fei and Wei, Haoran and others},
  journal={arXiv preprint arXiv:2412.15115},
  year={2024}
}

@article{bai2025qwen2,
  title={Qwen2.5-VL Technical Report},
  author={Bai, Shuai and Chen, Keqin and Liu, Xuejing and Wang, Jialin and Ge, Wenbin and Song, Sibo and Dang, Kai and Wang, Peng and Wang, Shijie and Tang, Jun and others},
  journal={arXiv preprint arXiv:2502.13923},
  year={2025}
}

\appendix

\section{Appendix}
\label{sec:appendix}
\subsection{Rendering Details}
\label{app:rendering_details}

The visual inputs used in our experiments were rendered programmatically from the same underlying structured instances as the textual inputs. We used task-specific rendering templates for each task in this paper.

\subsubsection{Matrix}

\begin{figure*}[htbp]
    \centering
    \includegraphics[width=0.9\linewidth]{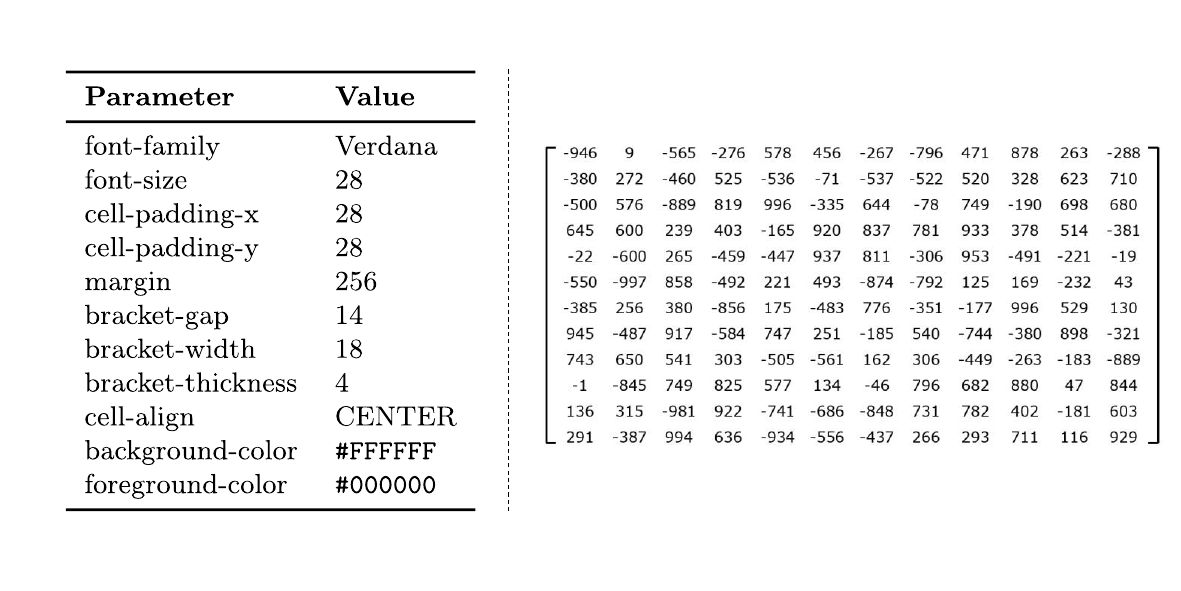}
    \caption{
    Rendering parameter setting for matrix visual inputs. The left column lists the values for each parameter, while the right column shows a representative rendered example. Images are rendered at their natural size without additional resizing or normalization.
    }
    \label{fig:transpose_render_params}
\end{figure*}

The font-family and font-size determine how entries are rendered, cell-padding-x and cell-padding-y set the horizontal and vertical spacing between entries, and margin defines the outer whitespace. The bracket-gap, bracket-width, and bracket-thickness specify the position and appearance of the enclosing brackets, while cell-align controls how values are positioned within each column. Background-color and foreground-color define the overall color scheme.

\subsubsection{Conway Grid}

\begin{figure*}[htbp]
    \centering
    \includegraphics[width=0.9\linewidth]{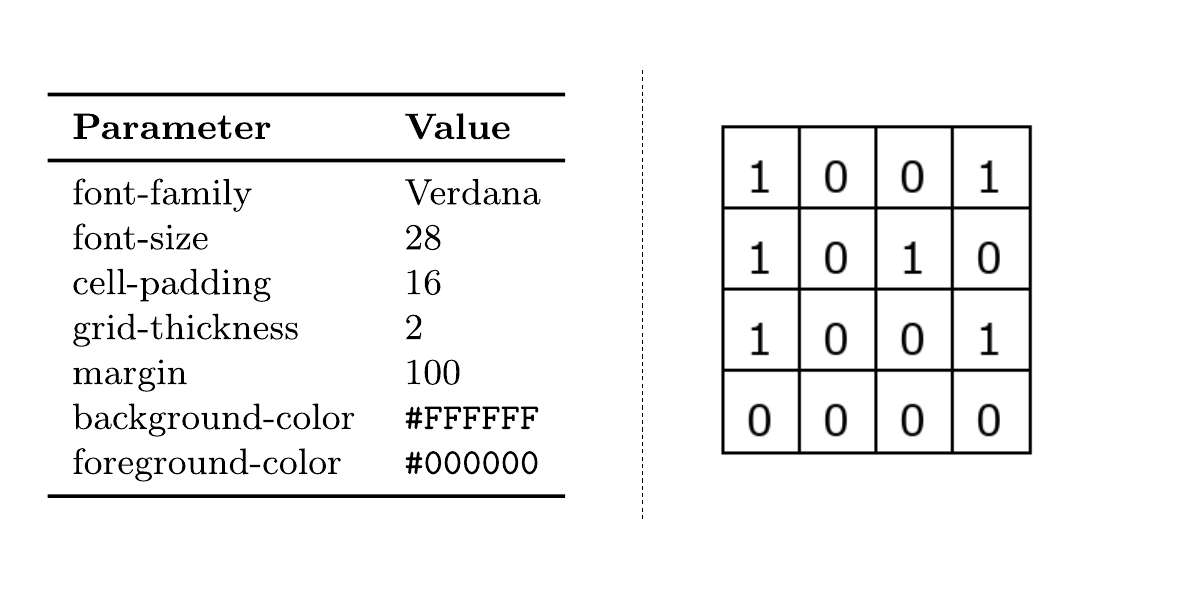}
    \caption{
    Rendering parameter setting for Conway grid visual inputs. The left column lists the values for each parameter, while the right column shows a representative rendered example. Images are rendered at their natural size without additional resizing or normalization.
    }
    \label{fig:conway_render_params}
\end{figure*}

The font-family and font-size control text appearance, cell-padding sets the internal spacing within each cell, and grid-thickness determines the visibility of row and column boundaries. Margin defines the outer whitespace, while background-color and foreground-color control the visual contrast.

\subsubsection{Disruptive Matrix}

\begin{figure*}[htbp]
    \centering
    \includegraphics[width=0.9\linewidth]{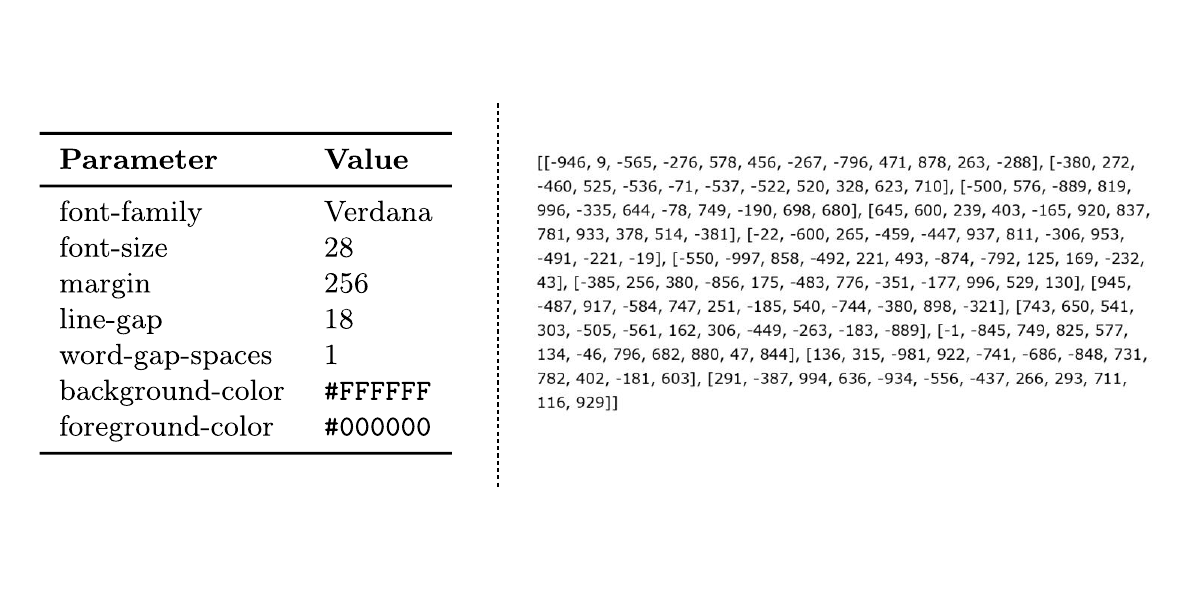}
    \caption{
Rendering parameter setting for disruptive matrix visual inputs. The left column lists the values for layout, font, spacing, while the right column shows a representative rendered example. Images are rendered at their natural size without additional resizing or normalization.
    }
    \label{fig:dis_render}
\end{figure*}

The font-family and font-size determine how tokens are rendered, while margin defines the outer whitespace of the canvas. Line-gap controls the vertical spacing between lines, and word-gap-spaces sets the number of spaces inserted between tokens, affecting horizontal density. Background-color and foreground-color specify the visual contrast.

We do not fix the text region width manually. Instead, we derive it from the original matrix layout by measuring rendered entries under the target font and accounting for inter-column spacing, brackets, and outer margins. The plainflow string is then rendered within this derived canvas, with wrapping controlled by the canvas geometry together with the word spacing and line spacing settings.

\subsection{Task Definitions}
\label{app:task_definitions}

We study three structured tasks: matrix transpose, Conway's Game of Life, and LU decomposition. In all cases, the underlying task is fixed across representations. The same problem instance may be presented either as serialized text or as a rendered 2D input, but the required transformation and correctness criteria remain unchanged.

\paragraph{Matrix Transpose.}
Given an input matrix $A$, the model is asked to output its transpose $A^\top$. The input is a square numeric matrix with integer entries. The expected output is the full transposed matrix as a 2D array. Correctness requires exact agreement with the ground-truth transpose.

\paragraph{Conway's Game of Life.}
Given the current board state of Conway's Game of Life, the model is asked to output the full next-generation board of the same size. The input is a binary grid in which 1 denotes a live cell and 0 denotes a dead cell; input cells are sampled with equal probability for the live and dead states. The next board is computed according to the standard Game of Life rules~\citep{gardner1970fantastic} under zero-padding boundary conditions: each cell is updated synchronously based on the number of live neighbors in its 8-cell neighborhood, a live cell remains live if and only if it has 2 or 3 live neighbors, and a dead cell becomes live if and only if it has exactly 3 live neighbors; otherwise, the cell is dead in the next generation. Cells outside the board are treated as dead. The expected output is the full next-generation grid as a 2D array of binary values, and correctness requires exact agreement at every cell position.

\paragraph{LU Decomposition.}
Given an input matrix $A$, the model is asked to produce matrices $L$ and $U$ such that $A = LU$, where $L$ is a lower triangular matrix and $U$ is an upper triangular matrix. The input is a square numeric matrix. We use LU decomposition without pivoting, i.e., matrices are constructed such that a valid factorization exists without row permutations. The expected output is the pair $(L, U)$ in structured matrix form. Correctness is defined functionally: the predicted factors must satisfy the triangular constraints and reconstruct the input matrix under matrix multiplication, up to a small numerical tolerance.

\subsection{Implementation Details}
\label{app:implementation_details}

\paragraph{Models.}
Across all tasks, we use the same two model families under two representation conditions. The 1D condition uses GLM-4-9B-0414~\citep{glm2024chatglm} with serialized textual inputs, while the 2D condition uses Glyph~\citep{cheng2025glyph}, which connects GLM with a vision encoder. This same GLM/Glyph pair is used for matrix transpose, Conway's Game of Life, and LU decomposition.

\paragraph{Data generation and splits.}
All datasets in this study are synthetic and programmatically generated. For each task, the 1D and 2D conditions are derived from the same underlying task instances, so that the comparison differs in representation rather than in problem content. Training and test instances are generated separately, and evaluation is always performed on held-out test sets.

For matrix transpose, each example consists of a square integer matrix together with its gold transpose. The standard single-size experiments cover matrix sizes from $12\times12$ to $20\times20$. For each size, data are split into train/test partitions with a ratio of 5:1. The mixed-training setting combines examples from $12\times12$, $14\times14$, and $16\times16$ matrices in a 1:1:1 ratio.

For Conway's Game of Life, each example consists of a binary grid together with its full next-generation grid under standard Game of Life dynamics. The standard single-size experiments cover grid sizes from $4\times4$ to $8\times8$. For each size, data are split into train/test partitions with a ratio of 5:1. The mixed-training setting combines examples from $4\times4$, $5\times5$, and $6\times6$ grids in a 1:1:1 ratio.

For LU decomposition, each example consists of a square integer matrix generated by multiplying a lower triangular matrix and an upper triangular matrix with integer entries from $-9$ to $9$. The standard single-size experiments cover matrix sizes from $3\times3$ to $6\times6$. For each size, data are split into train/test partitions with a ratio of 5:1. The mixed-training setting combines examples from $3\times3$, $4\times4$, and $5\times5$ matrices in a 1:1:1 ratio.

\paragraph{Input and output format.}
Across both modalities, the task instruction remains the same and only the representation of the structured input changes. In the 1D condition, the prompt contains the serialized matrix or grid directly in text. In the 2D condition, the same instruction is used, but the structured input is provided as a rendered image. In both conditions, the model is trained to produce a reasoning trace enclosed in \texttt{<think>...</think>} tags followed by the final structured answer. Evaluation is performed only on the extracted answer portion after the closing tag.

\paragraph{Evaluation.}
Task definitions and correctness criteria are specified in Appendix~\ref{app:task_definitions}. For matrix transpose and Conway's Game of Life, we report exact-match accuracy on the full output grid. For LU decomposition, we parse the predicted $L$ and $U$ and verify triangular constraints, and check that $LU$ reconstructs the input matrix within an absolute tolerance of $10^{-6}$.

\paragraph{SFT configuration for matrix transpose and Conway's Game of Life.}
For both matrix transpose and Conway's Game of Life, we use supervised fine-tuning with LoRA~\citep{hu2021lora} for both GLM and Glyph. Unless otherwise noted, the two models share the same optimization hyperparameters: 2 epochs, an initial learning rate of $5\times10^{-6}$ decayed to $2\times10^{-6}$ with cosine scheduling, a warmup ratio of 0.03, \texttt{bf16} precision, maximum sequence length 8192, LoRA rank 32, LoRA $\alpha=32$, and LoRA dropout 0.05. Effective batch size is controlled through per-device batch size and gradient accumulation. In our experiments, the effective batch size is fixed within each task: we use batch size 32 for matrix transpose and batch size 8 for Conway's Game of Life. These task-specific choices are shared across GLM and Glyph, so within each task the 1D and 2D conditions are trained under the same effective batch size.

\paragraph{RLVR configuration for LU decomposition.}
For LU decomposition, we use reinforcement learning with verifiable reward. Specifically, we adopt DAPO~\citep{yu2025dapo}, a variant of GRPO~\citep{shao2024deepseekmath}, implemented with the verl library~\citep{sheng2025hybridflow}. Unless otherwise noted, training uses a group size of 8, 16 rollouts per prompt, and a learning rate of $2\times10^{-6}$. The same RLVR setup is applied to both GLM and Glyph.

\subsection{LU Decomposition Case Study}
\label{app:lu_case_study}
\begin{figure*}[t]
    \centering
    \includegraphics[width=\linewidth]{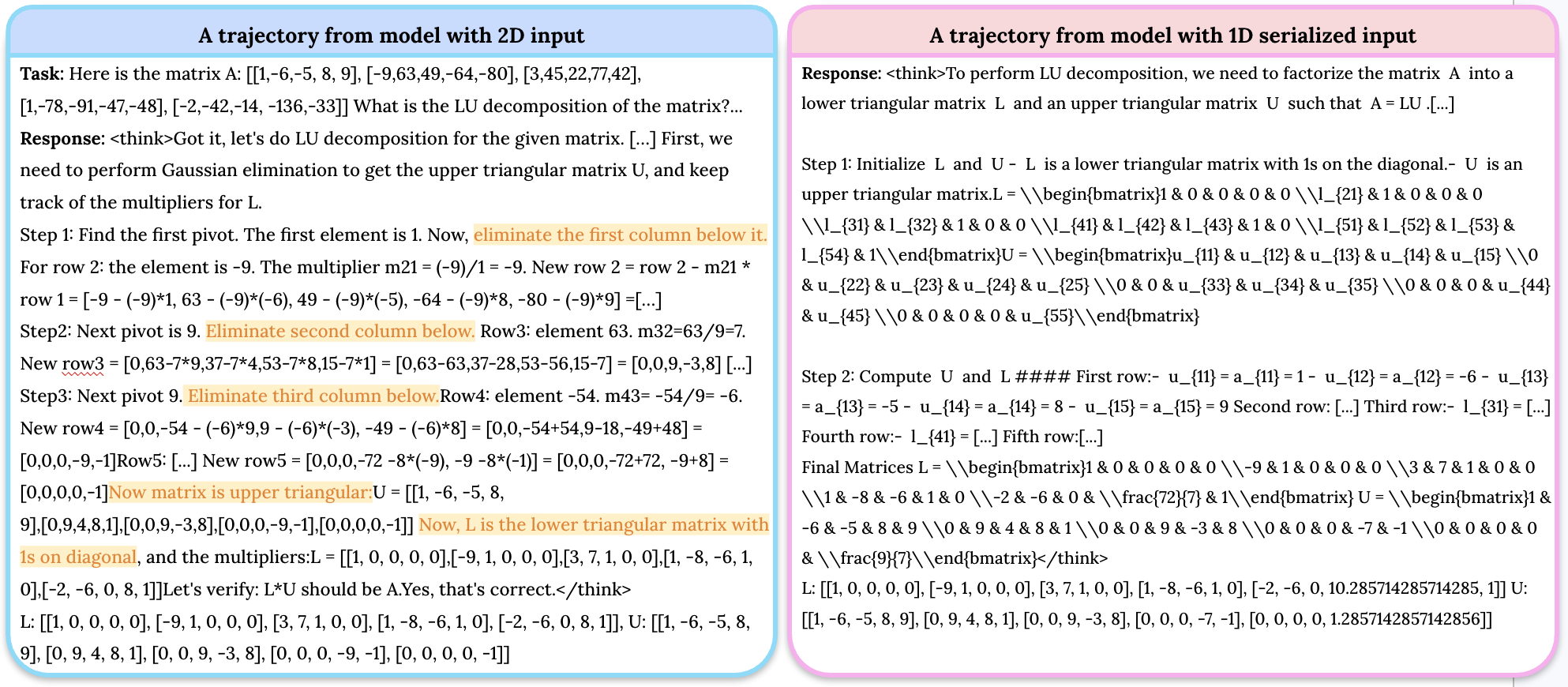}
   \caption{Representative reasoning trajectories for LU decomposition under 2D layout (left) and 1D serialization (right). Orange highlights in the 2D trajectory mark phrases where the model explicitly references the spatial structure of the rendered input to organize its elimination steps.}
    \label{fig:lu_case_study}
\end{figure*}

Figure~\ref{fig:lu_case_study} compares how the two models approach the same LU decomposition instance. Under 2D layout, the model structures its reasoning around the visible matrix, producing step-by-step elimination with explicit spatial references such as ''eliminate the first column below it.'' Under 1D serialization, the model attempts a similar procedure but relies on symbolic index tracking without spatial grounding, leading to less organized intermediate steps and numerical errors in the final output.

\subsection{Qwen Matrix-Transpose Experiment Details}
\label{app:qwen_transpose}

The experiment follows the matrix-transpose setup described in Appendix~\ref{app:implementation_details}. We use the same input formatting, supervised fine-tuning protocol, and exact-match evaluation criterion as in the main matrix-transpose experiments. Both models are trained on a mixed-size transpose dataset containing $4\times4$, $6\times6$, and $8\times8$ matrices, with the three sizes mixed in a $1{:}1{:}1$ ratio. Evaluation is then performed separately on each matrix size.

For the visual condition, matrices are rendered using the same rendering pipeline and parameter settings as the main matrix-transpose visual inputs, as described in Appendix~\ref{app:rendering_details} and illustrated in Figure~\ref{fig:transpose_render_params}. Because this Qwen comparison uses a different model scale and compares a base text-only model with an instruct-tuned vision-language model, we use it only as a supplementary qualitative check rather than as a controlled estimate of cross-family effect size.

\end{document}